\documentclass[sn-mathphys-num]{sn-jnl}
\usepackage{xcolor}
\usepackage{graphicx}%
\usepackage{multirow}%
\usepackage{amsmath,amssymb,amsfonts}%
\usepackage{amsthm}%
\usepackage{mathrsfs}%
\usepackage{todonotes}
\usepackage[title]{appendix}%
\usepackage{xcolor}%
\usepackage{textcomp}%
\usepackage{manyfoot}%
\usepackage{booktabs}%
\usepackage{algorithm}%
\usepackage{algorithmicx}%
\usepackage{algpseudocode}%
\usepackage{listings}%
\usepackage{tikz}
\usepackage{subcaption}

\newcommand{\ew}{\widetilde{w}}
\newcommand{\eA}{\widetilde{A}}

\newcommand{\CIT}{\textit{CIT}}
\newcommand{\PUB}{\textit{PUB}}

\begin{document}
\title{Explaining word embeddings with perfect fidelity: A case study in predicting research impact}
\author[1]{\fnm{Lucie} \sur{Dvo\v{r}\'a\v{c}kov\'a}}
\author[2]{\fnm{Marcin P.} \sur{Joachimiak}}
\author[1]{\fnm{Michal} \sur{\v{C}ern\'y}}
\author[3]{\fnm{Adriana} \sur{Kubecov\'a}}
\author[4]{\fnm{Vil\'em} \sur{Sklen\'ak}}
\author[4]{\fnm{Tom\'a\v{s}} \sur{Kliegr}}%\email{tomas.kliegr@vse.cz}

\affil[1]{Department of Econometrics, Prague University of Economics and Business, Czech Republic}
\affil[2]{Environmental Genomics and Systems Biology Division, Lawrence Berkeley National Laboratory,  USA}
\affil[3]{Department of System Analysis, Prague University of Economics and Business, Czech Republic}
\affil*[4]{ Department of Information and Knowledge Engineering,  Prague University of Economics and Business, Czech Republic}

\affil[*]{Correspondence: tomas.kliegr@vse.cz%, lucie.dvorackova@vse.cz
}

\abstract{
The best-performing approaches for scholarly document quality prediction are based on embedding models. In addition to their performance when used in classifiers, embedding models can also provide predictions even for words that were not contained in the labelled training data for the classification model, which is important in the context of the ever-evolving research terminology.
Although model-agnostic explanation methods, such as Local interpretable model-agnostic explanations, can be applied to explain machine learning classifiers trained on embedding models, these produce results with questionable correspondence to the model.
We introduce a new feature importance method, Self-model Rated Entities (SMER), for logistic regression-based classification models trained on word embeddings. We show that
SMER has theoretically perfect fidelity with the explained model, as the average of logits of SMER scores for individual words (SMER explanation) exactly corresponds to the logit of the prediction of the explained model. 
Quantitative and qualitative evaluation is performed through five diverse experiments conducted on 50,000 research articles (papers) from the CORD-19 corpus. Through an AOPC curve analysis, we experimentally demonstrate that SMER produces better explanations than LIME, SHAP and global tree surrogates. 
}


\maketitle

\section*{Introduction}
A recent influential study by Tshitoyan et al. pointed to the ability of word embeddings to predict new thermoelectric materials that were indeed reported in future years in material engineering journals \citep{tshitoyan_unsupervised_2019}. 

This demonstrates that embedding-based machine learning models can establish a significant relationship between certain concepts well in advance of them being recognized in a scientific field and beyond. We build on this property of word embeddings to capture latent meaning, extending the work of Tshitoyan et al., who relied on vector similarity to find novel, desirable chemical materials, to the supervised setting of building predictive machine learning models.

The emerging generation of methods for the prediction of the impact or quality of research builds upon the latest advances in text representation \cite{xue2023re,de2023multischubert,hirako2023realistic}.  
These algorithms are typically based on the projection of text elements into $\mathbb{R}^d$, so-called \emph{embeddings}.
Embeddings are real-valued vectors representing words, sentences, or whole documents that are supposed to encode semantic meaning. It is expected that word-, sentence- or document-vectors that are closer in $\mathbb{R}^d$ are similar in meaning. 

Word embeddings perform well in tasks related to the identification of impactful research. This included the task of citation prediction \cite{hirako2023realistic} as well as predicting whether a paper will be accepted or rejected \cite{de2023multischubert,xue2023re}. However, the evaluations in these approaches have focused largely on model predictive performance, while many applications require reliable explanations. Having the information that a draft of the paper will not be accepted is not as useful to the author as a specific and reliable explanation of what specifically contributes to its positive or negative perception by the model.
We previously investigated several state-of-the-art model-agnostic methods for explanations, including LIME and Shapley plots, in our previous work \citep{beranova_why_2022},
noting that the quality of the explanations generated for why specific research was cited lags behind white-box models such as rule learning. 
These methods often obscure the explanation process by introducing randomly generated data and additional machine-learning models. The quality of such explanations is not yet thoroughly studied, with research showing that methods such as LIME can be manipulated to have very low fidelity \citep{slack2020fooling}.

In this publication, we, therefore, propose a new explanation method that provides intrinsic explanations of embedding-based predictive models \emph{without} training additional models or generating new instances. Furthermore, we theoretically show that the newly proposed explanation method is guaranteed to have perfect fidelity for logistic regression predictive models. We empirically confirm this result in five experiments.

\section*{Method}
This section describes the workflow of our methodology for predicting the impactfulness of words as well as entire documents.  A reliable prediction of research impact for any research paper only based on its abstract is a difficult task. In our research, we \emph{address a narrower subproblem of predicting which research will be groundbreaking}. That is, we are interested in identifying the most influential research -- both papers and newly emerged entities, such as drugs or virus strains in the CORD-19 training data context. The focus on the most confident predictions makes the research impact and explainability problem more tractable.

We first describe our new method of building an embedding-based classifier for predicting paper impact, which will directly provide word-based explanations. Then, we will cover two existing baseline methods (ORC and LIME). Based on this, we will provide an overview of the framework for predicting the impact of research papers and impactful entities (words). Finally, we will cover the methods we used for evaluation in the five experiments.

\subsection*{Self Model Entities Rated (SMER)}\label{sect:smer}
We represent a scientific paper by its abstract, which is a string consisting of dictionary words. 
Let $\mathcal{W}$ stand for a \emph{dictionary}; an element $w \in \mathcal{W}$ is called a \emph{word}. A \emph{document}, a \emph{text}---or specifically in this study, an \emph{abstract} of a scientific paper---is a nonempty sequence $A = (w_1, \dots, w_N)$ of words.   We are given a family $\mathcal{A}$ of abstracts
$A_i = (w_{i,1}, \dots, w_{i,N_i})$, $i = 1, \dots, n$, called \emph{corpus}. 
We assume that $\mathcal{W} = \bigcup_{i=1}^n \bigcup_{j=1}^{N_i} w_{i,j}$.

Given a text corpus as a training set, the trained model assigns a \emph{score} to each text feature. 
Thus the score estimated by our model provides a quantitative measure of \emph{feature importance}.

\paragraph{Embedding of words}
We use the corpus $\mathcal{A}$ to produce an~\emph{embedding} $\ew \in \mathbb{R}^d$ for each word $w \in \mathcal{W}$. In this work, we assume the use of non-contextual embeddings, which represent the dictionary $\mathcal{W}$ as a family $\{\ew\in\mathbb{R}^d \mid\ w \in \mathcal{W}\}$ of points in $\mathbb{R}^d$ which can be subsequently used as inputs to other procedures requiring real-valued input data.

The most significant property of embeddings for this study is that \emph{when words $w, w' \in \mathcal{W}$ frequently appear ``close'' to each other in the training dataset $\mathcal{A}$, then $D(\ew, \ew')$ is ``small''}, where $D$ stands for some distance measure on $\mathbb{R}^d$, such as the cosine similarity or Euclidean distance.

  \paragraph{Embedding of documents} 

Having constructed an embedding $w \in \mathcal{W} \mapsto \widetilde w \in \mathbb{R}^d$ of words, the next step is
a construction of an \emph{embedding of abstracts}, which is a function $A \mapsto \widetilde A \in \mathbb{R}^d$, where $A$ is an abstract --- a sequence of words $A = (w_1, \dots, w_N).$ 
We use the \emph{average embedding}
\begin{equation}
\eA = \frac{1}{N}\sum_{j=1}^N \ew_j.
\label{eq:wax}
\end{equation}

{
This aggregation function has the following favorable property of \emph{linearity}: 
$\widetilde A$ is the average of the vectors 
$\widetilde w_1,\dots,\widetilde w_N$, a feature essential for the proofs of Propositions 1–3.
}

\paragraph{A classifier: Logistic regression} The next step is the training of a classification model. Assume that a target variable is available for supervised learning. That is, let our dataset consist of $(A_i, y_i)_{i=1, \dots, n}$, where $y_i \in \{0,1\}$ is a measure of a research impact of a paper with abstract $A_i$. 
In our work, values $\{0,1\}$ are used as indicators of whether the research is highly cited/impactful (coded as 1), or not (coded as 0). 

Logistic regression is used here as a natural representative of classifiers. Given that it is a linear additive model, it is often used as a base learner when an explanation is needed \cite{henrik}. Recall that logistic regression assumes the regression relationship 
\begin{equation}
\mathsf{E}[y_i\mid A_i] = 
L(\beta_0 + \beta^T \eA_i), \quad
i = 1, \dots, n, \label{eq:logiregbasic}
\end{equation}
with 
\begin{equation}
L(\xi) = \frac{1}{1 + e^{-\xi}}\label{eq:logifun}
\end{equation}
being the logistic function and $\beta_0 \in \mathbb{R}$, $\beta \in \mathbb{R}^d$ parameters. (Here, $\mathsf{E}[\,\cdot\!\mid\!\cdot\,]$ stands for the conditional expectation.) Training the model amounts to finding estimates $\widehat\beta_0, \widehat\beta$ of the model parameters $\beta_0, \beta$ e.g.~by minimization of the residual squares $\sum_{i=1}^n [y_i - L(\widehat\beta_0 + \widehat\beta^T \eA_i)]^2$ or by maximization of the log-likelihood 
$\sum_{i=1}^n\left[ y_i \log L(\widehat\beta_0 + \widehat\beta^T \eA_i) + (1-y_i) \log (1 - L(\widehat\beta_0 + \widehat\beta^T \eA_i))\right]$, often with restrictions on the norm $\|(\beta_0, \beta^T)\|$ (e.g.~by adding a constraint into the optimization 
or by adding a regularization penalty into the objective function).

\paragraph{Score of an abstract} Given an abstract $A$, we can predict its impactfulness%, called here \emph{score}, 
 using the trained logistic model by computing 
\begin{equation}
\textit{SCORE}(A) := L(\widehat\beta_0 + \widehat\beta^T\eA).
\label{eq:score}
\end{equation}

\vspace{\baselineskip}

\paragraph{SMER Score: Logistic regression  on word embeddings is self-explainable}
The basis of our self-explanatory embedding-based classification framework is the observation that a word $w \in \mathcal{W}$ can be seen as a single-word abstract $A = (w)$. We can thus compute feature importance (i.e. word importance) directly using the same model, which was trained to classify all documents (the SCORE from Eq.~\ref{eq:score}). When applied to a single word, the formula reduces to 
\begin{equation}
\textit{SCORE}(w) = L(\widehat\beta_0 + \widehat\beta^T\ew).\label{eq:myscore}
\end{equation}

Words $w \in \mathcal{W}$ can be sorted according to their scores; by selection of a cutoff threshold, we get a classifier of high- and low-scoring words, i.e., words that have high or low potential to contribute to the number of paper's citations. Since $\ew$ is an embedding vector, the method can also score words unseen in the training data as long as these can be assigned an embedding vector.

\subsection*{Properties of SMER}

\paragraph{Notation} For the sake of brevity, let us introduce abbreviations for the logits
$$
\lambda(w) := \widehat\beta_0 + \widehat\beta^T\ew, \quad 
\lambda(A) := \widehat\beta_0 + \widehat\beta^T\eA,
$$
where $w$ is a word and $A$ is an abstract. 

Recalling that $L$ is the logistic function (\ref{eq:logifun}),
we have $\textit{SCORE}(w) = L(\lambda(w))$,
$\textit{SCORE}(A) = L(\lambda(A))$. The following observation shows an important property.

\paragraph{Proposition 1 (linearity of logits)} \emph{If $A = (w_1, \dots, w_N)$ is an abstract, then
$$
\lambda(A) = \frac{1}{N}\sum_{\ell=1}^N \lambda(w_{\ell}).
$$}

\emph{Proof.} Using (\ref{eq:wax}) we can write
$$ \lambda(A) 
= \widehat\beta_0 + \widehat\beta^T\eA
= \widehat\beta_0 + \widehat\beta^T\left(\frac{1}{N}\sum_{\ell=1}^N\ew_{\ell}\right)
= \frac{1}{N}\sum_{\ell=1}^N \left(\widehat\beta_0 + \widehat\beta^T\ew_{\ell}\right)
= \frac{1}{N}\sum_{\ell=1}^N \lambda(w_{\ell}).\qed
$$

\paragraph{Proposition 1 is a key property for assessing the fidelity.}

{
\emph{Fidelity is the percentage of predictions that are the same between the
original model and the explanation} \cite{lakkaraju2017interpretable,markus2021role}.
Formally, for a finite evaluation set $\mathcal D \subset \mathcal A$:
\[
\mathrm{Fidelity}(g,f;\mathcal D)=
\frac{1}{|\mathcal D|} \sum_{A \in \mathcal D} \mathbf{1}\bigl[g(A) = f(A)\bigr].
\tag{F}\label{eq:fidelity}
\]
Here, \(f\colon \mathcal A \to \mathbb{R}\) is the prediction of the original model,
and \(g\colon \mathcal A \to \mathbb{R}\) is the prediction of the explanation method being evaluated.

In our case, fidelity holds \emph{in the logit space}, where
\[
f(A) = \lambda(A), \qquad
g(A) = \tfrac{1}{N} \sum_{j=1}^N \lambda(w_j),
\]
and Proposition~1 ensures
\[
f(A) = g(A) \quad \text{for all } A \in \mathcal D.
\]

For presentation, we apply the logistic function $L(\cdot)$:
\[
\textit{SCORE}(A) := L(f(A)), \qquad 
\textit{SCORE}(w_j) := L(\lambda(w_j)).
\]
This rescaling improves interpretability, and since $L$ is strictly increasing (Proposition~2), the ranking of words is preserved.
}

\paragraph{Proposition 2 (order preservation)} \emph{The ordering of dictionary words according to their \textit{SCORE}s
is the same as their ordering according to the logits ($\lambda(w)$-values).} 

\emph{Proof.} Since
$\textit{SCORE}(w) = L(\lambda(w))$ and
$L$ is an increasing function, for any pair of 
dictionary words $w,w'$ we have 
$\textit{SCORE}(w) \leq \textit{SCORE}(w')$ if and only if $\lambda(w) \leq \lambda(w')$. \qed

\vspace{.5cm}

Proposition~2 tells us that if we want to sort the dictionary words according to
their scores, we can equivalently sort them according to their logits ($\lambda(w)$-values). %It follows that the logit can be also used as a score measure.
However, in contrast to the scores, the logits do not possess the nice property of lying in the interval (0,1). 

\paragraph{Proposition 3 (monotonicity of \textit{SCORE}s)} 
\emph{If $A = (w_1, \dots, w_{k-1}, w_k, w_{k+1}, \dots, w_N)$ 
and 
$A' = (w_1, \dots, w_{k-1}, w'_k, w_{k+1}, \dots, w_N)$
are abstracts, then 
$$
\textit{SCORE}(w'_k) > \textit{SCORE}(w_k)\ \ \Longrightarrow\ \ \textit{SCORE}(A') > \textit{SCORE}(A).$$} 
 (In other words: if the score of a word in an abstract increases, then the score of the abstract increases as well.)

\emph{Proof.} $L$ is an increasing function; thus $\textit{SCORE}(w'_k) > \textit{SCORE}(w_k)$ implies $\Delta := \lambda(w'_k) - \lambda(w_k) > 0$.
Now  
\begin{align*}
\textit{SCORE}(A) 
  &= L\bigl(\lambda(A)\bigr) 
   = L\left(\frac{1}{N}\sum_{\ell=1}^N \lambda(w_\ell)\right)
   < L\left(\frac{1}{N}\sum_{\ell=1}^N \lambda(w_\ell) + \frac{\Delta}{N}\right)
\\&= L\left(\frac{1}{N}\!\left(\sum_{\ell \in \{1, \dots, N\}\setminus\{k\}} \lambda(w_{\ell})\right) + \frac{\lambda(w'_k)}{N} \right)
   = L\bigl(\lambda(A')\bigr) = \textit{SCORE}(A').
\end{align*}
The strict inequality again follows from the fact that $L$ is increasing.\qed

\vspace{.5cm}

\subsection*{Reference baseline feature importance: ORC and LIME}
In this section, we define two reference models.  One is based on training a logistic regression model over bag-of-words, which is the previous-generation approach for representing document content. This method is also intrinsically explainable as the importance of individual words follows from their assigned logistic regression coefficients. Finally, we cover LIME, a state-of-the-art technique commonly used to explain embedding-based classification models.  
\paragraph{BoW representation of abstracts} We continue with the notation from the previous section. 
Let the dictionary $\mathcal{W}$ consist of $M$ words $v_1, \dots, v_M$.
Bag-of-Words (BoW) represents an abstract $A$ by a vector $\bar A \in \mathbb{R}^M$ of dictionary words present in the abstract:
\begin{equation}
\label{eq:bow}    
\bar A_k = \left\{
\begin{array}{ll}
1, & \text{\ \ \ if $v_k \in A$,} \\
0  & \text{\ \ \ otherwise,}
\end{array}\right.
\quad \quad k = 1, \dots, M.
\end{equation}

\paragraph{Logistic~regression}
The corpus $\mathcal{A}$ is represented by vectors $\bar A_1, \dots, \bar A_n \in \mathbb{R}^M$ which serve as explanatory variables in the logistic regression model
$$
\mathsf{E}[y_i\mid A_i] = L(\gamma_0 + \gamma^T \bar A_i), \quad i = 1, \dots, n,
$$
with coefficients $\gamma_0 \in \mathbb{R}$ and $\gamma \in \mathbb{R}^M$.
The estimate of $\gamma$ is denoted by $\widehat\gamma$. 

\paragraph{ORC: Odds ratio change as a measure of word importance in BoW models}
Observe that $\widehat \gamma_k > 0$ indicates that \emph{adding} the word $v_k$ into an abstract (if not yet present) increases the odds of the paper to become highly cited. On the other hand, $\widehat \gamma_k < 0$ indicates that adding $v_k$ into an abstract is predicted to decrease the odds of the paper becoming highly cited. Therefore, the coefficient $\widehat\gamma_k$ measures the impact of the dictionary word $v_k$ on the target variable $y$. Recall that within the logistic regression framework 
this effect can be interpreted in terms of odds ratio. We define 
\begin{equation}
\textit{ORC}(v_k) := e^{\widehat\gamma_k}, \quad k = 1, \dots, M.
\label{eq:ORRC}
\end{equation}
For BoW, entity (word) importance thus directly corresponds to feature importance as measured by Odds Ratio Change (ORC), because features are binary word indicators.

\paragraph{LIME: Locally Interpretable Model Explanations}\label{sect:BoW}

The LIME method (Local Interpretable Model-agnostic Explanations) \citep{ribeiro2016why} is a technique for explaining model decisions at the level of individual data instances. This method focuses on creating an interpretable model that approximates the behavior of the original model in the vicinity of a specific data instance.

The explanation formulated by the LIME method for a word \( v_k \) is expressed as:

\[
E(v_k) = L(f, g, \Pi_{v_k}) + \Omega(g), 
\]

where \( E(v_k) \) represents the explanation for the word \( v_k \), \( L(f, g, \Pi_{v_k}) \) denotes the measurement of unfaithfulness of the model \( g \) when approximating \( f \) in the locality \( \Pi_{v_k} \), and \( \Omega(g) \) represents the complexity of the model \( g \).

\subsection*{Framework for predicting impactful research}
\begin{figure}[h]
    \centering
    \includegraphics[width=\textwidth]{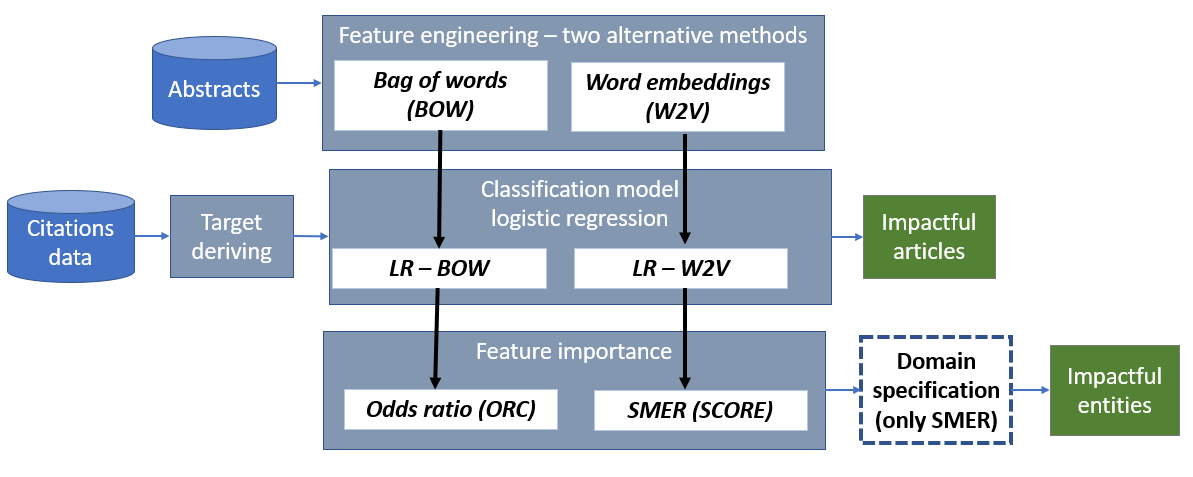}
    \caption{The figure shows the methodological pipeline of workflow steps.}
    \label{fig:schema3}
\end{figure}

Our empirical study employs the methodology from Figure~\ref{fig:schema3}. Here, 
we cover the technical details of the phases depicted in the figure:
feature engineering, target derivation, training of the classification model, estimating feature importance, model validation,
and domain specification.
%\todo[inline]{In Fig 1 address overlaps of text}

\paragraph{Data sources}
\label{sec:data_prep}
The data utilized for this study is derived from the CORD-19 corpus  \citep{wang2020cord}, which consists primarily of research papers pertaining to biology and medicine relevant to SARS-CoV-2 and that were published between 1952-2022. Specifically, we employed the version of the corpus downloaded on April 23, 2022, comprising a total of 992,921 papers. 
Paper citation counts were extracted from  \url{Opencitations.org}. 
For the training of classifiers, we used only paper abstracts from the years 2016-2020 for which the Opencitation data were available.                          
                                               
\paragraph{Preprocessing}
The number of papers was reduced to 598,587 after filtering out abstracts with duplicate document object identifiers (doi). Subsequently, papers whose number of characters did not exceed 10 were removed, reducing the paper count to 483,781 papers. Filtering of non-English abstracts detected using the pre-trained 
Fasttext \citep{
joulin2016bag,joulin2016fasttext,cavdar_fasttext-langdetect_nodate} language identification 
model reduced the count to 477,237 papers. For training embedding models, only papers with recent years of publication (2016-2021) were used and for training classification models, 80\% of all papers with citation data for the specified years of publications were used.
The total number of unique abstracts we used for training embeddings was 225,679. Note that to train word embedding models, more papers were available as abstracts with missing citation data could also be used.

The abstracts were preprocessed as follows. Punctuation was removed except for dashes (as these are used in compound biomedical names), all text was converted to lowercase, and single numbers were deleted. Lemmatization was used to convert the words into root words. After this preprocessing, papers with less than 20 characters were removed. The final dataset contained 476,175 papers. After preprocessing, we also checked for possible duplications, determining that more than 99.8\% of abstracts were unique.

To train embeddings, we used the Skip-Gram version of Word2vec, which is based on a prediction of the context of a given word. Although the alternative CBoW models are known for their speed, the Skip-Gram model is recommended for datasets with infrequent words \cite{mikolov2013distributed}, which corresponds to our focus on emerging research concepts. The skip-gram model was already used for this purpose in related work \citep{tshitoyan_unsupervised_2019}.
 
\paragraph{Predictors}
After the preprocessing steps, paper abstracts were represented using two variant methods: first, embedding-based word vectors generated from abstracts with~Eq.~\ref{eq:wax}, and second, we used an intrinsically explainable model based on bag-of-words (BoW) (see~Eq.~\ref{eq:bow}).

However, unlike several recent approaches,
we deliberately excluded the author information and other metadata such as journal impact factor, from training, since we have found in our earlier work that it can lead to publication biases \citep{beranova_why_2022}.

The  \emph{dimension} $d \geq 1$ is a parameter used to train the embedding models.  For training the word embeddings, we used embedding vector dimensions of 100 and a window size of 10. 
These settings were based on our prior experience combined with recommendations from the literature \citep{mikolov13}.

\paragraph{Citation count as a proxy for research significance}
Citation counts are a commonly used proxy for research impact obtainable on a large scale but with some limitations. For example, recent authoritative research on this topic \cite{rodriguez2021total} states:  \emph{"it should be strongly emphasized that citation counts correlate with the scientific relevance or impact of a scientific publication, but they do not always measure the relevance of a specific scientific publication".}

This shows that while citations often fail to reflect the true impact for various reasons,  this often happens on an individual level, which is not relevant to our work as our machine learning model learns an abstraction. Our research is based on this widely held assumption, well summarized by \cite{aksnes2019citations}: 
\emph{"Although most citation analysts seem to agree that citing or referencing is biased, it has been argued that this bias is not fatal for the use of citation as performance indicators—to a certain extent, the biases are averaged out at aggregated levels ... the presence of different cognitive meanings of citations and motivations for citing does not necessarily invalidate the use of citations as (imperfect) performance measures...”}

An example of a well-documented bias is that in biosciences,  positive (statistically significant) results are cited more frequently \cite{duyx2017scientific}. This is a potentially favourable bias for our research, as we expect impactful new concepts or research to be mostly supported by statistically significant evidence.

\paragraph{Citation target construction}\label{sect:yidef}
To define the target, we combine two subtasks of scholarly document quality prediction: predicting citation counts and predicting whether a paper will be rejected or accepted \cite{de2023multischubert}. 
We define the target variable $y_i$ as a 0-1 indicator of
being a \emph{highly cited paper}.
To define the notion precisely, let
\begin{align}
\CIT(i,t) &:= \text{number of citations of paper $A_i$ originated in year $t$}, \label{eq:CIT} \\
\PUB(i)   &:= \text{year of publication of paper $A_i$.} \label{eq:PUB}
\end{align} 
For example, $\CIT(i,\PUB(i))$ is the number of citations of $A_i$ from its publication year. 
By definition, if $t < \PUB(i)$, then $\CIT(i,t) = 0$. The number $\sum_{t=\PUB(i)}^{2022} \CIT(i,t)$ represents the citation count of $A_i$ from its publication year up to the moment of our download of the citation data.

\paragraph{Balancing class distributions}\label{sect:balanc}
Recall that $n$ stands for the number of corpus papers. Both training and evaluation are performed on distinct time slices defined by the set $T$ of \emph{publications years}, e.g.~$T = \{2016, 2017\}$
and a fixed \emph{citation year} $\tau$, e.g.~$\tau = 2018$. Then, paper $A_i$ with $\PUB(i) \in T$ is defined to be \emph{highly cited} with respect to $T$ and $\tau$ if the number of its citations from year $\tau$ (strictly) exceeds the median:
\begin{align}
y_i &= \left\{
\begin{array}{ll}
1, & \text{if $\CIT(i,\tau) > \text{median}_{1 \leq k \leq n}\{\CIT(k, \tau) \mid \PUB(k) \in T\}$,} \\
0  & \text{otherwise,} \\   
\end{array}
\right. \nonumber \\
& \hspace{5cm}\mbox{where } i \in \{\iota \mid 1 \leq \iota \leq n, \PUB(\iota) \in T\}.\label{eq:yi}
\end{align}

Table~\ref{tab:data} presents the composition of the low- and highly cited papers in the dataset after the target has been derived. The count of papers that were published in a given year differs in individual citation years due to missing citation data. These were not consistently available for each paper in all covered years. Some papers are cited before they have been published --- this situation can occur at the preprint stage of the paper.
Following the strict inequality in definition~(\ref{eq:yi}), the two subsets separated by the median are not completely equally sized. For example, assume a corpus containing five papers with citation counts $0, 1, 1, 1, 2$, then as a result of the preprocessing, there will be only one paper in the  ``high'' category, i.e.~paper with the citation count \emph{strictly} above the median~$1$, and the remaining four papers with at most one citation will be assigned to the ``low'' category.

\begin{table}[h]\centering
\rotatebox{0}{\resizebox{\textwidth}{!}{\begin{tabular}{@{}ccccccccccccc@{}}
\toprule
\multicolumn{1}{l}{}                                                                   & \multicolumn{12}{c}{\textbf{Citation Year}}                                                                                                                                                                           \\
                                                                                       & \multicolumn{2}{c}{\textbf{2017}} & \multicolumn{2}{c}{\textbf{2018}} & \multicolumn{2}{c}{\textbf{2019}} & \multicolumn{2}{c}{\textbf{2020}} & \multicolumn{2}{c}{\textbf{2021}} & \multicolumn{2}{c}{\textbf{2022}} \\
\multirow{-2}{*}{\textbf{\begin{tabular}[c]{@{}c@{}}Publication \\ Year\end{tabular}}} & \text{low}    & \text{high}   & \text{low}    & \text{high}   & \text{low}    & \text{high}   & \text{low}    & \text{high}   & \text{low}    & \text{high}   & \text{low}    & \text{high}   \\ \midrule

%\textbf{2014}                                                                          & 742             & 646             & 767             & 610             & 736             & 620             & 762             & 663             & 253             & 122             & 0               & 0               \\
%\textbf{2015}                                                                          & 887             & 742             & 876             & 735             & 840             & 767             & 901             & 774             & 289             & 109             & 1               & 0               \\

\textbf{2016}                                                                          & 962             & 629             & 903             & 799             & 867             & 861             & 918             & 871             & 305             & 157             & 2               & 0               \\
\textbf{2017}                                                                          & 375             & 368             & 979             & 616             & 894             & 874             & 1,057            & 797             & 341             & 139             & 1               & 1               \\
 
\textbf{2018}                                                                          & 13              & 0               & 403             & 397             & 1,020            & 771             & 1,040            & 997             & 350             & 167             & 1               & 0               \\
\textbf{2019}                                                                          & 2               & 1               & 13              & 1               & 532             & 497             & 1,284            & 1,248            & 441             & 192             & 0               & 0               \\

\textbf{2020}                                                                          & 9               & 0               & 5               & 0               & 120             & 6               & 11,955           & 8,727            & 5,204            & 3,458            & 11              & 0               \\
\textbf{2021}                                                                          & 0               & 0               & 0               & 0               & 0               & 0               & 53              & 29              & 33              & 10              & 0               & 0               \\

\hline\textbf{Total}                                                                         & 2,990            & 2,386            & 3,946            & 3,158            & 5,009            & 4,396            & 17,970           & 14,106           & 7,216            & 4,354            & 15              & 1               \\ \bottomrule
\end{tabular}}}
\caption{Number of low- and highly cited papers by publication and citation year in the used subset of the CORD-19 corpus used for training and evaluation of classification models. 
}
\label{tab:data}
\end{table}

\paragraph{{Justification of Binary Target Selection}}
{
We also explored a regression-based approach, predicting the exact number of citations and applying a threshold to derive class labels. However, as demonstrated in our previous work \citep{beranova_why_2022}, this approach resulted in lower predictive performance compared to the direct binary classification setup. Consequently, we adopted a classification formulation (with a binary instead of continuous target). It should be emphasised that even with a binary target (high vs low cited categories), it is still possible to discern the most significant research, since this is reflected in the probability of the classification. An article predicted 99\% to be in the high category is considered more significant than an article with 51\%  predicted probability.   In this way, the binary target formulation provides both more accurate and interpretable means for identifying impactful research compared to predicting the number of citations directly.}

\section*{Experiment 1: Predictions of impactful papers}\label{sect:qe}

The goal of this experiment is to evaluate the predictive performance on the task of identifying the most (and least) promising research. We compare the proposed method, logistic regression over word embeddings (further referred to as Word2vec or w2v), with an older approach based on BoW. 

\paragraph{Datasets}

For models using embeddings, for each of these datasets, embeddings were trained separately on abstracts published in the corresponding years. This experiment evaluates four consecutive two-year windows (2016–2017 to 2019–2020). Each window supplies \emph{all} abstracts for training word embeddings, see number of abstracts in Table~\ref{tab:exp1_data} 

\begin{table}[h]
\centering
\begin{tabular}{@{}lcc@{}}
\toprule
\textbf{Publication window} & \textbf{Word embeddings} & \textbf{Classification} \\
\midrule
2016–2017 & 12\,589 & 1\,867 \\
2017–2018 & 13\,036 & 1\,916 \\
2018–2019 & 14\,503 & 2\,256 \\
2019–2020 & 132\,422 & 18\,571 \\
\bottomrule
\end{tabular}
\caption{Number of abstracts used in Experiment 1.}
\label{tab:exp1_data}
\end{table}

To train the classification models, we created four datasets, each covering papers published in two consecutive years from the period 2016-2020. 
 The overview of the datasets is given in Table~\ref{tab:first_results}.

\paragraph{Evaluation strategies}

The time-based nature of the data requires a specific approach to data splitting. We report results using the time-split and random-split techniques, two commonly used approaches. Since the last year needs to be used for testing, models trained under time-split testing are not trained on the latest papers, which negatively affects their ability to evaluate the impact of recent research. We therefore also report results using a third technique, which addresses this issue, which we call semi-split testing.

\begin{description}

\item[\textbf{Random-split testing.}]
Random, time-sensitive splits of data are a common evaluation technique. In our setup, 80\% of the data from selected periods is used to train the models and the remaining 20\% from the same periods is used for testing.

\item[\textbf{Time-split testing.}]
This split is stricter than random-split testing, as the testing set consists of papers that were both published and cited later than the training set in order to prevent any leakage from the test data to the training data.

For training, we use 80\% of available papers from publication years $t-1, t-2, \dots$ with the target being derived from citation counts of these papers accumulated by year $t-1$. The test papers were those published in year $t$ with the target being derived from citation counts also from year $t$. This is a particularly challenging setup because the prediction of citations for papers published in the same year is often very difficult.

\item[\textbf{Semi-split testing.}]
Since the last year needs to be used for testing, models trained under time-split testing are not trained on the latest papers, which negatively affects their ability to evaluate the impact of recent research.
This semi-split evaluation technique best corresponds to how we envisage that the models would be used by leveraging the latest available citation counts for testing. {
Let \(t\) be the most recent publication year. For training, we take 80\% papers from \(t-1\) to \(t\), using citation counts accumulated up to the end of year \(t\). The remaining 20\% of the \(t\)-cohort form the test set and are evaluated by the citations they collect during the following calendar year \(t+1\). Thus, the model learns from the freshest vocabulary while still being judged on genuine one-year-ahead citation predictions.}

\end{description}

\paragraph{Focus on groundbreaking research}
To evaluate the ability of the model to separate groundbreaking research from uninfluential papers, we construct a special testing set \emph{Extreme10\%}. Using citation count as a proxy for the impact of the research. This is a subset of the respective test set containing a selection of 5\% of papers with the highest citation counts and 5\% of papers with the lowest citation counts.  

\paragraph{Evaluation measures for the prediction task} %exp1
The Area Under the ROC Curve (AUC) is used as the main metric to evaluate predictive performance. %, known also as ROC's Gini coefficient. 
Since the target is binary and both classes are balanced, the values of AUC are almost identical to the values of accuracy and F1-Score, which are thus not reported.

\paragraph{Evaluation scenarios}
The last column of Table~{\ref{tab:first_results}} identifies the given combination of publication and citation year used in the train and test data and type of split as the evaluation scenario. For example, for training the classification model in scenario 1 (publication years 2016 and 2017), 80\% of all papers published in 2016+2017 with citations in 2017 were used, which makes a total of (962+375+629+368) = 2334 x 0.8 = 1867 papers, which also corresponds to the value in Table \ref{tab:exp1_data}.

\subsection*{Results}

The results are presented in Table~\ref{tab:first_results}.

\begin{table}[h]
\resizebox{\textwidth}{!}{\begin{tabular}{@{}cc|cc|lccl@{}}
\toprule
\multicolumn{2}{c|}{\bf Training   set}                        & \multicolumn{2}{c|}{\bf Testing set} & \textbf{Split Type} & \multicolumn{2}{c}{\bf AUC\,--\,Extreme 10\% of Testing set}     &  \\ \midrule
\textsf{Publication year}            & \textsf{Cit. year}         & \textsf{Publication year} & \textsf{Cit. year} & & \multicolumn{1}{c}{\textsf{BoW}} & \multicolumn{1}{c}{\!\!\!\!\!\!\!\!\textsf{W2v}} & \\ \midrule
\multirow{3}{*}{80\% subset 2016, 2017} & \multirow{3}{*}{2017} & 20\% subset 2016, 2017       & 2017  & random-split         & 0.60                    & \bf 0.71                 & \{1\}   \\
                                   &                       & all 2016, 2017          & 2017  & random-split         & \bf 0.98                    & 0.81                  & \{2\} \\
                                   &                       & { \textit{all 2018}}                    & { \textit{2018}}  & time-split         & { \textit{0.76}}                    & {\bf \textit{0.80}}                  &{\{3\}}  \\ \midrule
\multirow{3}{*}{80\% subset 2017, 2018} & \multirow{3}{*}{2018} & 20\% subset 2017, 2018       & 2018  & random-split       & \bf 0.79                    & 0.78                & \{4\}    \\
                                   &                       & all 2017, 2018          & 2018  & random-split       & \bf 0.94                    & 0.80               & \{5\}     \\
                                   &                       & { \textit{all 2019}}                    & { \textit{2019}}  & time-split       & { \textit{0.59}}                    & {\bf \textit{0.67}}           & \{6\}         \\ \midrule
\multirow{3}{*}{80\% subset 2018, 2019} & \multirow{3}{*}{2019} & 20\% subset 2018, 2019       & 2019  & random-split    & 0.75                    & \bf 0.76            & \{7\}        \\
                                   &                       & all 2018, 2019          & 2019  & random-split    & \bf 0.96                    & 0.75           & \{8\}         \\
                                   &                       & { \textit{all 2020}}                    & { \textit{2020}}  & time-split    & {\bf \textit{0.52}}                    & { \textit{0.47}}            & {\{9\}}       \\ \midrule
\multirow{3}{*}{80\% subset 2019, 2020} & \multirow{3}{*}{2020} & 20\% subset 2019, 2020       & 2020  & random-split         & 0.65                    & \bf 0.83   & \{10\}                  \\
                                   &                       & all 2019, 2020          & 2020  & random-split         & \bf 0.93                    & 0.83& \{11\}                    \\
                                   &                       & \textit{20\% subset 2020}                    & \textit{2021}  & semi-split    & \textit{0.73}                    & \textbf{\textit{0.77}}   & \textit{\{12\}}     \\ \bottomrule
\end{tabular}}
\caption{Quantitative validation of LR-W2v and LR-BoW models. The best result for each evaluation setup is in bold. The time-split/semi-split is considered as most representative (in italics). }
\label{tab:first_results}
\end{table}

Although in terms of random-split evaluation, the BoW method most often has a higher AUC %($\{2\}$, $\{4\}$, $\{5\}$, $\{8\}$, $\{9\}$, $\{11\}$)
, LR-W2v model outperforms LR-BoW in most of the time-split \& semi-split scenarios ($\{3\}$,$\{6\}$,$\{12\}$). The probable reason why in scenario $\bf\{9\}$, both models are failing as AUC is close to 0.5, is that the training occurred on pre-pandemic data and papers published during the pandemic were used for testing. The interest of researchers (and consequently the citations) shifted considerably as a result of the pandemic.

On the reference evaluation setup $\{12\}$, which was evaluated based on the most recent set of citations, the best AUC of 0.77 was obtained by the embedding-based model (trained with Word2vec). 
Overall, the embedding model seems to be better able to adapt to concept drift as measured in the time-split scenarios, where pre-pandemic data were used for training and evaluated on papers that originated subsequently. 

\section*{Experiment 2: Correlation of feature importance with actual impact}
\label{ss:exp2}

This is the first of two experiments aimed at evaluating the quality of feature importance ratings provided by SMER and comparing them against a baseline. As the baseline method in this experiment, we use coefficients from logistic regression over bag-of-words. The second experiment (Experiment 3) will present the comparison against LIME.

In the BoW mode, coefficients for individual features ($\textit{ORC}(w)$ of a word $w \in \mathcal{W}$ from Eq.~\ref{eq:ORRC}) directly correspond to word importance. As a second method, we represent the documents with uninterpretable features (word embeddings) and also train a logistic regression classifier. We then use the SMER method ($\textit{SCORE}(w)$) to identify specific words responsible for the classification. As was shown in Proposition 1, this explanation has also 100\% fidelity as the average of predictions for individual word vectors is equal to the prediction of the explained model.

The models were trained with and without regularization.  We chose the regularized versions (ridge regression) since they performed better in terms of correlation with \emph{ActImpact} for both models. 

\paragraph{Datasets}

The logistic regression was trained on papers published in 2019-2020 for which citation data were available. This experiment reuses the same corpus for scenario \textbf{2019–2020} already shown in Table~\ref{tab:exp1_data}.

The overview of the dataset is in  Table~\ref{tab:first_results} (scenario \#12). The citation counts from 2021 were used for the validation.

\paragraph{Evaluation strategy}

The explanations are evaluated through the correlation of feature importance with actual impact. 

A key element of our research is being able to identify which words or entities (drugs, vaccines, etc.) are linked to high research impact. 
For a well-performing measure of feature importance, we should be able to observe a positive correlation between feature importance and the presence of the word among papers with observed research impact. High positive feature importance (for the highly cited class) will correlate with the presence of the word among highly cited papers, and vice versa for negative feature importance.

With this intuition, we define the  actual impact of the word $w$  as
\begin{equation}
\textit{ActImpact}(w) = \frac{\#\{i \in \{1, \dots, n\} \mid w \in A_i,\ i \in H_{2021}\}}
{\#\{i \in \{1, \dots, n\} \mid w \in A_i,\ \PUB(i) \in \{2019, 2020\}\}},\label{eq:actimpact}
\end{equation}

where $\#$ denotes the number of elements of a set and $H_{2021}$ is the set of highly-cited papers in 2021 computed as follows:

\begin{align}
H_{2021} &= \{ i \in \{1, \dots, n\} \mid PUB(i) \in \{2019, 2020\}, \nonumber \\ 
&\hspace{0.5cm}\,\ \CIT(i,2021) > \textit{median}_{1 \leq k \leq n}\{\CIT(k, 2021) \mid \PUB(k) \in \{2019, 2020\}\}\}, 
\label{eq:H2020}
 \nonumber
\end{align}
where functions $\CIT$ and $\PUB$ were defined in Eq.~(\ref{eq:CIT},\ref{eq:PUB}). 

Given a trained model, we will use the top 100 words
with the highest \textit{SCORE}s (for LR-W2v) or \textit{ORC}s (for LR-BoW) and bottom 100 words with the lowest values and compare these with the actual impact according to (\ref{eq:actimpact}).
The baseline for this comparison is the probability of a paper being highly cited, computed as:
\begin{equation}
\label{eq:baseline}
p_{high} = \frac{\# H_{2021}}{\#\{i\ \mid\ \PUB(i) \in \{2019,2020\}\}}.
\end{equation}

\subsection*{Results}
\begin{figure}[h]        
        \includegraphics[width=5cm]{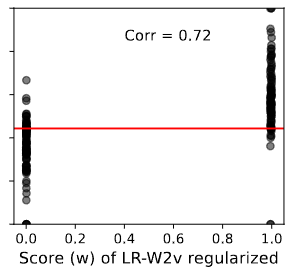} 
        \includegraphics[width=4cm]{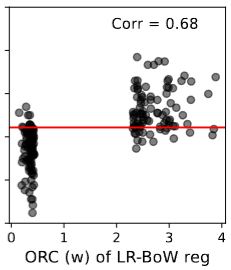}       
        \centering
        \caption{This figure shows the evaluation of feature importance values for SMER and a baseline method using the Pearson correlation coefficient between the Actual impact of the word and its predicted feature importance. 
        Left: The horizontal axis shows the score of a regularized Logistic Regression model trained on Word2vec embeddings (LR-W2V reg). Right: Reference method based on regularized Logistic Regression Odds Ratio Coefficients for BoW (LR-BoW reg).
        Vertical axis indicates the proportion of the papers containing a given word which belongs to the highly cited category. 
        Dots correspond to words $w \in T_{100}^{W2V} \cup L_{100}^{W2V}$ (left) and $w \in T_{100}^{BOW} \cup L_{100}^{BOW}$ (right figure). The red line corresponds to a random baseline ($p_{high}\approx 0.57$).
        }
        \label{fig:twin2}
    \end{figure}

Figure~\ref{fig:twin2} answers the question of whether the top 100 words with the highest predicted values, denoted as $T_{100}^{W2V}$ and $T_{100}^{BOW}$, frequently appear in papers that later turned out to be highly cited and whether words with the lowest predicted values, denoted as $L_{100}^{BOW}$ and $L_{100}^{W2V}$,  appear in abstracts of papers that later turned out to be lowly cited.

The SMER method has a higher correlation coefficient with the ground-truth word importance  (ActualImpact) than the baseline ORC method. This shows that even though the underlying classification model uses uninterpretable features, the SMER method was able to identify specific words responsible for the classification. Note that SMER results are even better than for the ORC method, which uses a representation where individual features directly correspond to word importance.  

\section*{Experiment 3: Comparing SMER through feature perturbations}

The previous two experiments showed that word embeddings are superior to the bag-of-words representation both in terms of predictive performance and, when coupled with the proposed SMER method, also when it comes to explanations. The method that we used in Experiment 2 to measure the quality of explanations was based on correlation with actual impact. In this next experiment, we compare SMER with LIME, SHAP and Global Surrogate Importance methods. 

\paragraph{Datasets}

The data utilized for training the embeddings and the logistic regression model were the same as detailed in Experiment 2, which comprised papers published in years 2019 and 2020, with citation counts from 2021.
For the application of both LIME and SMER explanations, we used the logistic regression model trained in Experiment 2.

\paragraph{Evaluation strategy} 

To compare the two feature importance methods used in our work, we used a standard approach based on Area Over the Perturbation Curve (AOPC) evaluation \citep{samek2016evaluating}. Specifically, we employed AOPCglobal \citep{vanderlinden2019global}, which is a modification of AOPC for benchmarking global feature importance measures. AOPCglobal, unlike the original AOPC, which assesses importance based on local explanations, operates by sequentially removing words based on their global importance. 
\footnote{Although the name refers to "Area Over the Perturbation Curve", the metric is commonly computed as the area \emph{under} this curve—specifically, under the plot of performance degradation as informative features are removed. This interpretation allows for intuitive comparison: higher area means the model relies more heavily on the top-ranked features, indicating that the importance ranking captures genuinely influential words.}

The SMER method readily provides global importance scores based on its internal structure, which can be directly used to rank and remove words for AOPCglobal evaluation.

To obtain these for LIME,  which is a local method, we utilized GALE (Global Aggregations of Local Explanations) \citep{vanderlinden2019global}. We used two variations of GALE: Global LIME Importance and Global Average LIME Importance \citep{vanderlinden2019global}.{
SHAP Importance was obtained by computing local SHAP values using the SHAP TextExplainer \citep{lundberg2017unified} and then aggregating the absolute contributions of each word across the test set, following the approach for estimating global importance from local explanations. 

In addition, we included \textit{Global Surrogate Importance}, obtained from a decision tree. The probabilistic output of the base model (a value in the 0–1 interval representing the probability of “high impact”) is treated as a \textit{continuous} target, so we train a regression decision tree on it using TF--IDF (ngrams of length 1--2, vocabulary limited to 5000 terms) inputs to approximate the probability outputs of the embedding-based classifier
The surrogate model was trained with a maximum depth of 100 and achieved a high fidelity of $R^2 = 0.89$ on the test set. This approach follows the model extraction framework of \citet{bastani2017interpreting}, which builds interpretable approximations of complex classifiers for post-hoc explanation purposes.

Next, features were removed in the order of global importance. The removals were cumulative, first feature, first two features, etc. After each removal, the model was applied to the test set (with removed words), and the average change (typically a decrease) in the predicted class probabilities was recorded. Based on this, the AOPCglobal curve was constructed by plotting the number of removed words (X-axis) against the decrease in the class probabilities (Y-axis). %The same approach was applied to SMER, using its feature importance scores directly, to obtain the AOPCglobal curve. 

To quantify the aggregate loss in predictive accuracy consequent to the systematic removal of the most significant features, we employed a trapezoidal rule integration, which was used to compute the \emph{area under the AOPCglobal curves}.

The curves were divided into equidistant segments, and areas of individual trapezoids were summed. For the evaluation of global metrics, the curve was segmented into 10 parts, corresponding to each curve consisting of a total of 11 data points (equivalent to the removal of 0 to 10 words).

\subsection*{Results}

LIME explainer was configured to use 15 features (with 10 being the default). As a result, the top 15 most influential words were selected for generating explanations for each abstract.

We also tried to perform minor adjustments in kernel width or the number of samples, but since this did not significantly alter the outcomes, other parameters were set to their default values.

\begin{figure}[h]
    \centering
    \includegraphics[width=\textwidth]{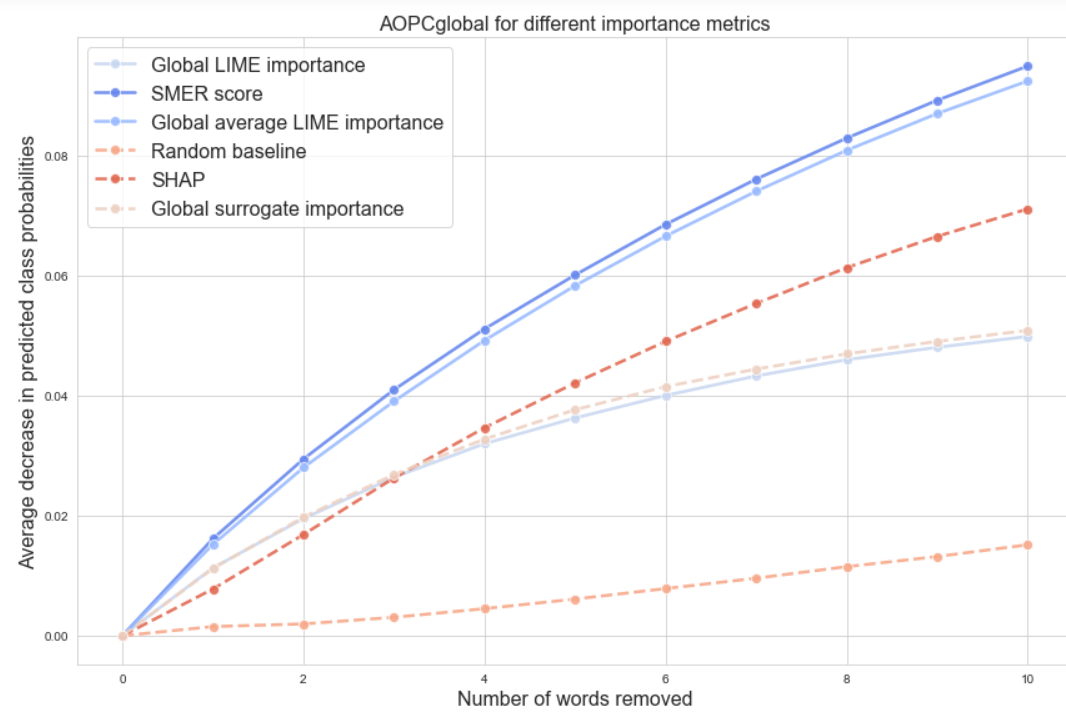}
    \caption{This figure presents the AOPCglobal for different feature importance metrics.
    The higher the decrease in predictive performance when top-ranked words are removed (from left to right), the better the feature importance method is in identifying salient features.
    }
    \label{fig:aopc}
\end{figure}

Figure~\ref{fig:aopc} shows a summary plot of the AOPCglobal comparison results for LIME, SMER, and a baseline that removed randomly selected words. In addition to these methods, we also evaluated SHAP Importance and Global Surrogate Importance.

SHAP Importance was obtained by aggregating local SHAP values across the test set. For each word, we computed the average of its absolute SHAP contributions, resulting in a global importance score.

Global Surrogate Importance was extracted from the trained decision tree by measuring the total reduction in prediction error (mean squared error) attributed to each feature across all splits. These values were then normalized and ranked.

Table~\ref{tab:auc_aopc} shows the area under the AOPC curve. The random selection, shown by a dashed line, remains almost constant at zero, showing that random word removal has no significant effect on the model's predictions, as expected. An increasing number of removed words results in decreasing changes in the average predicted class probability for all tested methods, which indicates that words associated with the highest scores according to these methods, which were removed first, had the biggest influence. Out of the two aggregation methods for LIME, Global Average LIME Importance had better results; nevertheless, the best overall results were obtained by SMER. 
SHAP Importance outperformed both Global LIME and Global Surrogate methods, although it did not reach the level of SMER or Global Average LIME Importance.

To further validate that this finding extends to other datasets, we compared the performance of SMER and LIME using the same AOPC-based methodology on the IMDB movie sentiment dataset unrelated to the biomedical domain. This evaluation showed the same pattern, with SMER achieving even better results in terms of the area under the AOPC curve compared to LIME. A link to this analysis and results is included in the Availability of code and data section.

\begin{table}[]
\begin{tabular}{@{}ll@{}}
\toprule
\textbf{Global metric}                     & \textbf{Area under the AOPCglobal curve} \\ \midrule
SMER SCORE                                 & 0.50                                     \\
Global Average LIME Importance             & 0.48                                     \\
{SHAP importance}          & {0.33}                   \\
{Global surrogate importance} & {0.27}               \\
Global LIME Importance                     & 0.26                                     \\
Random baseline                            & 0                                     \\ \bottomrule
\end{tabular}
\centering
\caption{Area under the AOPCglobal curve for global feature importance metrics. A higher number indicates a better result.}
\label{tab:auc_aopc}
\end{table}

\section*{Experiment 4: Qualitative evaluation of top-predicted biological entities}\label{ss:exp3}

The previous three experiments validated SMER through the quantitative evaluation of aggregate results using standard methodologies: train-test splits (experiment 1) and feature importance metrics (experiments 2 and 3).
The goal of this experiment is to validate SMER by evaluating the utility of \emph{specific predictions} of significant recently introduced virus strains, drugs and vaccines. {This analysis and interpretation were conducted with significant input and validation from our domain expert in biosciences (M.P.J.) to ensure the biological relevance.}

\paragraph{Datasets}

To train the model on the most recent data,  we used 6,964 papers published between 2019 and 2020 as the set of training instances for the logistic regression model. The citation counts were used from the year 2021. 

Word vectors (embeddings) were trained using the embedding-based model on recent papers published in the years 2019-2022 with a total of 414,174 papers. The rationale for including the year 2022 in the training of the embeddings was to simulate the situation where the text of papers is available but not their actual impact (citation counts).

\paragraph{Evaluation strategy}

The goal of the application of SMER in biosciences is to identify top candidates for important concepts, such as drugs and vaccines. The importance of a word $w$ is reflected in the value of $SCORE(w)$.
To have one compound measure reflecting the degree of impact (\emph{SCORE}) as well as the size of impact (\emph{NumArt}), we introduce the \emph{customized score}:

\begin{equation}
    \textit{CScore}(w) =  \textit{SCORE}(w) \times \textit{NumArt}(w), 
    \label{eq:cscore}
\end{equation}
where \textit{NumArt}(w) is the number of papers where the word appears:
\begin{equation}
    \textit{NumArt}(w) =  \#\{i \in \{1, \dots, n\} \mid w \in  A_i,\  2020 \leq \textit{PUB}(i) \leq 2022\}.
    \label{eq:numart}
\end{equation}

\subsection*{Results}

Table~\ref{tab:res1} shows the top 15 words sorted by CScore from words that did not appear before the year 2021. These words can be divided into three groups:  1)~variants of the SARS-CoV-2 virus, 2)~vaccines, and 3)~medicines and drugs.

\setlength{\tabcolsep}{5pt}
\begin{table}[!htb]
\centering
\begin{tabular}{r l l r r r}
\toprule
Rank & Word & Topic & SCORE & NumArt & CScore \\
\midrule
1  & omicron         & virus     & 0.998 & 1253 & 1250 \\
2  & b1351           & virus     & 0.972 &  981 &  954 \\
3  & b16172          & virus     & 0.998 &  714 &  713 \\
4  & {\bf b11529}    & virus     & 0.999 &  349 &  349 \\
5  & vitt            & condition & 0.844 &  205 &  173 \\
6  & b1617           & virus     & 0.998 &  122 &  122 \\
7  & casirivimab     & drug      & 0.996 &   92 &   92 \\
8  & {\bf covaxin}   & drug      & 0.958 &   95 &   91 \\
9  & b16171          & virus     & 0.992 &   91 &   90 \\
10 & {\bf imdevimab} & drug      & 0.997 &   86 &   86 \\
11 & pango           & virus     & 0.994 &   84 &   83 \\
12 & covishield      & vaccine   & 0.615 &  130 &   80 \\
13 & t478k           & mutation  & 0.988 &   73 &   72 \\
14 & b1526           & virus     & 1.000 &   69 &   69 \\
15 & e484q           & mutation  & 0.991 &   69 &   68 \\
\bottomrule
\end{tabular}
\caption{Top 15 words sorted by $CScore = \text{SCORE} \times \text{NumArt}$, where SCORE is the predicted impact using SMER (Eq.~\ref{eq:myscore}) and NumArt (Eq.~\ref{eq:numart}) is the count of papers containing the word. Topic is a manually assigned semantic category. Words in bold were selected for a follow-up analysis in Table~\ref{tbl:b11cov}.}
\label{tab:res1}
\end{table}

From Table~\ref{tab:res1}, representatives of each of the three topics were selected by the highest score value and used as seeds to retrieve additional highly-scored words in each domain.  
We selected all words whose word-vector similarity (cosine $\geq$ 0.70) to the seed exceeded the threshold, sorted them by \textit{SCORE}, and kept the top 15 hits (excluding the seed itself).

In addition to the three tables generated automatically by selecting the highest ranked words in each category, we generated several additional tables, which are located in the supplementary material: 
{
\begin{itemize}
  \item Table~\ref{tab:b11529} — virus variants related to the seed “b11529”
  \item Table~\ref{tab:covaxin} — vaccine-related terms similar to “covaxin”
  \item Table~\ref{tab:imdevimab} — antibody and drug terms similar to “imdevimab”
  \item Table~\ref{tab:remdesivir} — drug terms similar to “remdesivir”
\end{itemize}}
Of note is the  fourth table (Table~\ref{tab:remdesivir}) containing the highest predicted words by similarity to  ``Remdesivir''. This choice of the seed word followed a previous analysis \citep{dornick_analysis_2021} that detected topics related to COVID-19 and identified the antiviral drug Remdesivir as the dominant term for the drug topic. The word ``Remdesivir'' on its own has with $\textit{SCORE}(w)=0.961$, as shown in~Table~\ref{tab:remdesivir},  also a high value of the predicted significance, although not as high as for imdevimab ($\textit{SCORE}(w)=0.997$), which was thus chosen as the representative for the drug topic.

\section*{Experiment 5: Qualitative evaluation of top-predicted papers
}

Unlike Experiment 4, which focuses on predictions of entities, this one focuses on the prediction of impactful papers. The predictions are validated through the bibliometric profile of the publishing journals and authors. 

\paragraph{Datasets}
We analyzed ``young'' papers published from the year 2021 which have the greatest potential (Score) according to the model.This experiment is based on the same data and predictive model as the previous experiment.

\paragraph{Evaluation strategy}

The predictions of impactful papers are evaluated through bibliometric indicators. 

Bibliometric indicators are widely accepted to correlate with research quality. For example, combined impact factor and citation counts were previously used to predict research impact \cite{levitt2011combined}. 
As a supplementary approach for validating predictions, we use bibliometric information related to the journal where the paper appeared, and a bibliometric profile of its authors. We emphasize that this information has not been used in our training setup and is used only for validation.

The quality of the journal is measured through the ISI impact factor, which is computed as the citation counts to papers the journal published in 2019-2020 from papers published in the year 2021 in other journals with an impact factor divided by the number of papers published by the journal in the years 2019-2020. As a reference for a high-impact prediction, we use the median impact factor among the top 10\% journals in the  ``Health sciences'' category (OECD category FORD 3030, data for the year 2021). 
To represent the research reputation of the authors, we use the maximum of the first and last authors' Hirsch-index (H-index) values. The H-index is defined as the  number of papers published by the author that have at least as many citations as this number. As a reference for a high-impact prediction, we use the median H-index of full professors, which was determined by analyzing the medical education journal editorial boards \citep{doja2014h}. Previous research has found that the H-index was a good indicator of citation in the first few years after publication \cite{levitt2011combined}.

We also include additional impact metrics from social media as retrieved from \url{Altmetric.com} in September 2022:  the number of tweets mentioning the research and the aggregate attention received on social media (such as the number of Facebook likes)  weighted by the relative reach of each type of source (Facebook, YouTube, Twitter).

\subsection*{Results}

Figure \ref{fig:graph_impact_factor} analyses a subset of papers with the highest predicted impact (Score), which were published in journals with an impact factor.
Among the most recent papers (published in 2022), the paper that appeared in a journal assigned the highest impact factor value (39.2) is also predicted to have the highest impact according to our method (SMER score). Out of the eight papers with an impact factor, seven appeared in journals assigned an impact factor higher than the median of 4.74 in the source corpus.  Note that the impact factor or other journal metadata were not part of the training corpus. Similarly, the authors of most top-ranked papers included an author with an H-index exceeding the median for full professors on journal editorial boards.
The high impact factors of journals in which the predicted impactful papers were published, as well as the track record of their authors, confirm the ability of our model to identify future research impact.  
 The Digital Object Identifiers (DOIs) of the specific top-predicted papers are available in the supplementary material.

\begin{figure}[h]
    \centering
    \includegraphics[width=350pt]{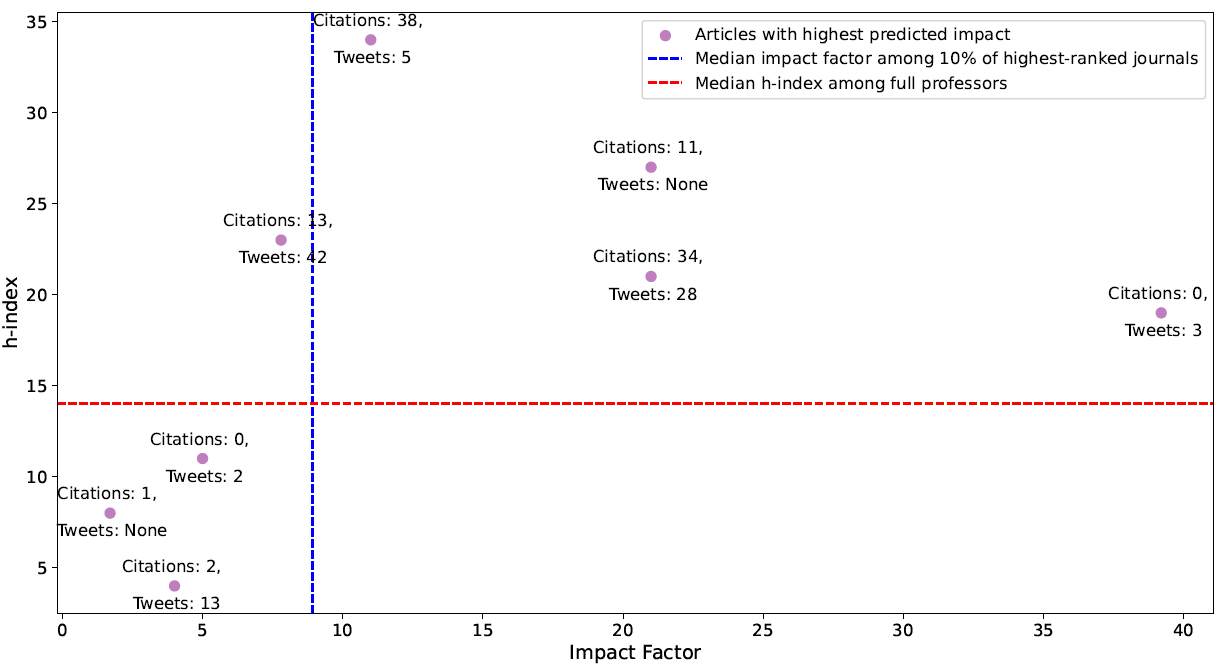}
    \caption{This figure shows the top papers by predicted impact (Score) published in 2021. 
    The reference impact factor (blue line) represents the median among the top 10\% journals in the category ``Health sciences'' category. %There are 1024 journals, we took first 10% by impact factor and 
    The reference H-index (red line)  corresponds to the median H-index of full professors. The attention score and tweets were retrieved from Altmetric.com. 
    }
    \label{fig:graph_impact_factor}
\end{figure}

\section*{Discussion}\label{sect:discussion}
%\subsection*{4.1 Analysis of predicted impactful entities} 

In this section, we present a more detailed examination and discussion of the predicted impactful entities from Experiment 4. {This analysis and interpretation were conducted with significant input and validation from our domain expert in biosciences (M.P.J.) to ensure the biological relevance.}

The inspection of the first year of appearance shows that in Table~\ref{tab:b11529}, there are a few words that predate the beginning of the COVID-19 pandemic in the year 2019. For example, ``Delta'' is a generic word, which has additional specific meaning in the COVID-19 context where it denotes the name of the virus mutation.  Similarly,  ``ba1'' is used to denote a variant of the SARS-CoV-2 virus from the year 2021. For the year 2020, the word ``ba1'' occurred just once as an antibody unrelated to COVID-19. The oldest word is ``monoclonal'', which appeared first in the year 1982 (in the studied corpus). Despite its relative age, it is predicted to have a high impact, possibly related to the fact that monoclonal antibodies are currently (as of 2024) one of the most successful treatment options for COVID-19.
 We note that all words in this table correspond to specific SARS-CoV-2 variants, with one exception being 'vocs'. This word corresponds to 'variants of concern' and represents a more general term for variants. Additionally, '202012/01' corresponds to another variant of concern 'VOC 202012/01' \cite{loconsole2021rapid}.
 
An important property of embeddings is that they can provide predictions even for words that were not contained in the labelled training data for the classification model. This is not possible with models based on the BoW representation. An example is ``nirmatrevil'', which has a high predicted score (0.99)  according to our proposed embedding-based logistic regression model. Since this word only appeared for the first time in a paper published in the year 2022, it was not included in the abstracts of any of the papers on which the logistic regression model was trained. 

In Table~\ref{tab:imdevimab}, we show results for the most important drug-related entities which are similar to the COVID-19 drug imdevimab. To summarize these findings, we note that of the top hits to this drug, seven of these top words are other COVID-19 antibody drugs with the antibody-drug designated suffix \textit{-mab}. Potentially other methods designed for learning word subtokens could identify further impactful patterns based on word prefix and suffix patterns. However, we observe that the identified important features similar to imdevimab are perfectly categorized as antibody drugs, combinations, or more abstract related terms ('monoclonal'), and include many synonyms. Thus included are: \textit{regen-cov} (representing the combination of \textit{casirivimab} (\#1) and \textit{imdevimab}, \textit{ly-cov555} (synonym for \textit{bamlanivimab} (\#3)), \textit{regn-cov2} (synonym for \textit{casirivimab-imdevimab}), \textit{ronapreve} (synonym for \textit{casirivimab-imdevimab}), \textit{ly-cov016} (also known as \textit{etesevimab} (\#2)), and \textit{regn10987} (synonym for \textit{imdevimab}). In total, including identified synonyms, 8 out of the top important and similar words to \textit{imdevimab} are synonyms of this drug or its combinations (e.g.~the combinations \textit{casirivimab-imdevimab}, \textit{regn-cov2} and \textit{ronapreve}). The remaining words among the top hits are other COVID-19 antibody drugs, with \textit{monoclonal} representing the general concept of monoclonal antibodies. With the exception of this general term, all other important words first appeared in the literature in 2020 or later, further validating their relevance for similarity to \textit{imdevimab}, which is a new drug developed for COVID-19.

In Table~\ref{tab:remdesivir}, we also considered similar and important features related to the drug \textit{remdesivir}, which has been used as a diagnostic word to assess word similarities for COVID-19 literature. In this case, all of the top important features were small-molecule COVID-19 drugs, with one exception being the antibody drug sarilumab approved for rheumatoid arthritis and repurposed for COVID-19. We note again that the suffix \textit{-ir}, indicating an antiviral compound, is found for 8 of the 14 most important words similar to \textit{remdesivir}, suggesting that subtoken approaches may provide additional benefits. Included in these top hits are some synonyms, e.g.~\textit{molnupiravir} and \textit{eidd-2801} (a preclinical name for this drug), \textit{lopinavir-ritonavir} as well as \textit{lopinavir/ritonavir} and \textit{lpv/r}. There are 11/14 instances of first mentions in the literature prior to 2020, all representing drugs that were repurposed for COVID-19. Thus, we note that with our methodology, it is also possible to automatically identify instances of drug repurposing from the literature.

\section*{Limitations}

In many applications, it is useful for features to correspond to multiword sequences such as bigrams or trigrams. Naturally, by defining $A := (w,w')$ for an entity-pair $w,w'$, we can
compute $\textit{SCORE}(A)$ using~(\ref{eq:score}). Sorting the entities according to their scores can reveal which \emph{connections of words} have a high potential to predict the high citability of a paper. Processing $n$-grams can have significance, especially in scientific areas such as medicine or materials engineering that often use multi-word expressions to denote concepts.

The area of multiword sequences is linked to the use of Word2vec method to generate embeddings in our experiments. The newer alternative approaches, such as BERT \citep{devlin2018bert},  generate different word embeddings for each context. 
The use of contextual embeddings might improve predictive performance, but the adoption of stable word embeddings empowers our method to deliver simple explanations.

{
Future work aimed at possible adaptations of our method for the use with contextual embeddings is outlined in the next subsection. Another possible source of performance improvements is the use of non-linear classifiers instead of logistic regression used in our experiments. 
In the last part of the limitations section, we discuss the possible benefits and disadvantages of using non-linear classifiers.}

\subsection*{Incorporating contextual word embeddings into SMER}

{
We have so far assumed a \emph{context-free} embedding, i.e.\ a fixed mapping  
\(w \in \mathcal{W} \longmapsto \widetilde w \in \mathbb{R}^{d}\).  
In contrast, a \emph{context-sensitive} (or \emph{contextual}) embedding allows the vector representation of a word to vary depending on the context in which it occurs.  

More formally, let \(A_1, \dots, A_n\) be the training abstracts, and for each dictionary word \(w \in \mathcal{W}\), define the set
\[
F_w := \left\{ (i, j)\ \middle|\ \text{the } j\text{-th word of } A_i \text{ is } w \right\}.
\]
For each \((i,j) \in F_w\), we define a context-sensitive embedding \(\widetilde w(i,j) \in \mathbb{R}^d\).  
That is, \(\widetilde w(i,j)\) may differ depending on the document \(A_i\) and position \(j\), even when referring to the same word \(w\).  
To emphasize this dependence on context, we write \(\widetilde w(i,j)\).

The main question is: How can context-sensitive embeddings be incorporated into the SMER methodology?  
There are at least two straightforward approaches.

\paragraph{Approach A: occurrence-average embedding}

We can simply construct a \emph{universal embedding} \(\widetilde w\) by averaging the context-sensitive embeddings over all occurrences of \(w\):
\[
\widetilde w := \frac{1}{|F_w|} \sum_{(i,j) \in F_w} \widetilde w(i,j), \quad w \in \mathcal{W},
\]
and use the resulting vectors as input to the SMER method in the same way as with context-free embeddings.

\paragraph{Approach B: contextualized scoring via singleton abstraction}

For each training abstract \(A_i = (w_{i1}, \dots, w_{i,N_i})\), we use the context-sensitive embeddings of its words at their respective positions \((i,j)\), and compute the abstract embedding as
\[
\widetilde A_i := \frac{1}{N_i} \sum_{j=1}^{N_i} \widetilde{w}_{ij}(i,j), \quad i = 1, \dots, n.
\]
These document embeddings \(\widetilde A_1, \dots, \widetilde A_n\) are then used to train the logistic regression model~(\ref{eq:logiregbasic}) to estimate \(\widehat \beta_0, \widehat \beta\). The resulting coefficients are global—they do not depend on any particular document or position.

To assign a score to a specific word in a fixed context, we construct an artificial one-word abstract \(A_{n+1} := (w)\), compute the corresponding context-sensitive embedding \(\widetilde w(n{+}1,1)\), and define its score in this minimal context as
\[
\textit{SCORE}(w) := L(\widehat\beta_0 + \widehat\beta^T \widetilde w(n{+}1,1)).
\]
This yields a score for each dictionary word in a consistent, minimal context, enabling a universal ranking across \(\mathcal{W}\).

\paragraph{Theoretical correspondence and limitations}

Approach A is, in essence, a natural extension of the original SMER method: it produces a single fixed embedding for each word by averaging its contextualized vectors, and then applies the standard SMER pipeline without further modification. As such, all theoretical properties established for SMER—most notably Proposition 1 (linearity of logits) and its consequences—remain valid under Approach A.

In contrast, Approach B operates with context-sensitive embeddings throughout, including during scoring. While it preserves the overall logic of SMER (computing document-level scores via logistic regression on average embeddings), the change in representation introduces subtle differences. Since the embeddings used at training and at scoring time may stem from different contexts, the theoretical properties of SMER are no longer automatically guaranteed. In particular, Propositions 1–3 have not been formally established under this setup and should therefore be interpreted with caution.
}

\subsection*{Extending SMER to nonlinear classifiers}

The embedding-based approach to predicting research impact achieves AUC-Extreme10\% values of 0.77 for the reference setup. While an AUC in the range between 0.7 and 0.8 is considered as \emph{acceptable} \cite{hosmer}, this does not mean that our method can, with acceptable reliability, assess the research impact of any paper. Instead, our research was aimed at evaluating the ability of machine learning models to identify the most promising research. {Enhanced performance could be obtained by the use of non-linear classifiers. However, this may result in the loss of the viable interpretability properties as discussed in the following.}

{
The SMER methodology fundamentally relies on the linearity of the decision function in the embedding space. Specifically, it assumes that the prediction for a document is based on a linear combination of its word embeddings—an assumption that enables key analytical properties such as the decomposition of document-level logits into word-level contributions (Proposition~1).

In practice, however, nonlinear classifiers such as neural networks or decision trees are commonly used to model more complex relationships in the data. A natural question arises: can the SMER scoring approach be meaningfully extended to such models?

One possible direction is to retain the general idea of SMER: scoring individual words by presenting them in a fixed, minimal context (e.g., a singleton abstract), while using a nonlinear classifier instead of a linear one. This would yield a score \(\textit{SCORE}(w)\) that reflects the model's prediction for that minimal input, even though the model no longer admits a linear decomposition.

However, this generalization comes with limitations. Without a linear decision surface, the prediction for a document can no longer be expressed as an average over its parts. Proposition~1 no longer holds, and fidelity cannot be guaranteed. 

Nonetheless, applying SMER-style ideas in the nonlinear setting may still be useful in exploratory or diagnostic contexts, particularly when interpretability is desired but strict decomposability is not required. Such extensions may enable broader use of the SMER framework in modern NLP pipelines, but they require careful validation.
}

\section*{Conclusions}\label{sect:conclusions}
This paper presents two interconnected contributions. Our original aim was to devise a supervised framework for predicting which research will become impactful. However, when we started working on this, we did not find a classification approach that would be both embedding-based and directly explainable. 
As the first contribution, we describe a new feature importance algorithm SMER that can explain a supervised logistic regression model trained on an embedding-based representation by applying the model to documents corresponding to embedding word vectors for single words. We first theoretically proved that this explanation approach has 100\% fidelity with respect to the underlying classification model.
 Next, we experimentally demonstrated the superiority of embedding-based models over the bag-of-words representation in terms of predictive performance and the quality of explanations if word importance is calculated with SMER. In another experiment, we showed that SMER produces a better global feature importance ranking than would follow from averaged LIME scores.

These experiments have shown the merits of the method quantitatively, but does it also generate useful predictions? Our second contribution is a new methodology for identifying specific newly published but potentially highly impactful entities as well as complete research papers.
We provide a detailed analysis of the significance of the top-predicted drugs, vaccines and viruses. The predictions are also validated through bibliometric analysis, which shows that the method correctly identified research written in some of the best journals by leading scientists.
Note that while most scientometric research relies on bibliometric factors such as journal quality, number of authors, etc., our training methodology was based solely on the content of the papers as reflected in the paper abstracts, and the bibliometric information is used purely for validation purposes. Therefore our method is less susceptible to introducing institutional, gender, and other biases into the machine learning models. 

\section*{Declarations}
\subsection*{Availability of code and data}
We used the implementation of logistic regression from \texttt{scikit-learn} package \citep{scikit-learn}.
As the text data, we used the CORD-19 corpus \cite{wang2020cord}.
The citations dataset is available from the authors at \url{https://github.com/beranoval/PRECOG/blob/main/citationcounts_oci_revised_year.zip}.\\
Code is available from the authors at \url{https://github.com/beranoval/PRECOG/tree/main}. 

\subsection*{Competing interests}
The authors have no potential conflicts of interest to disclose.

\subsection*{Funding} The work of M.~\v{C}ern\'y was supported by Czech Science Foundation under project 25-18028S.  The work of L. Beranov\'a Dvo\v{r}\'a\v{c}kov\'a was supported by grant IGA 16/2022 ``PRECOG: Predicting REsearch COncepts of siGnificance''. M.P. Joachimiak was supported by the Director, Office of Science, Office of Basic Energy Sciences, of the US Department of Energy, Contract No. DE-AC0205CH11231.  T.~Kliegr was supported by the long-term support for research activities of FIS VSE and the EU under agreement No 857446 (HeartBit 4.0). The authors would like to thank Karel Douda for acquiring the citation counts. 

\subsection*{Author contributions}
L.D. data processing and analysis, V.S. journal quality data and analysis, A.K. comparison between LIME and SMER and AOPC analysis, {L.D. adaptation of AOPC analysis for SHAP and global surrogates}, M.P.J.  interpretation of results {from the biosciences perspective}, T.K. and L.D. writing, M.\v{C}. formalization, A.K.
L.D. and T.K. methodology, T.K. conceptualization, approval of the final manuscript, all authors.

\bibliography{bibliography.bib}

\clearpage
\appendix
\section*{Supplementary Material}
\setcounter{table}{0}
\renewcommand{\thetable}{S\arabic{table}}

An interesting observation is that Table~\ref{tab:covaxin} contains several terms related to the Moderna vaccine (``moderna'', ``spikevax'', ``modernas'', ``mRNA-1273''). The two main terms from this list (by frequency) are  ``moderna'' and ``mRNA-1273'', both with a count of about 800 and very similar importance (SCORE) of 0.9548 and 0.9681, respectively.
The high agreement in the score for these close synonyms, which, however, have distinct word vectors and were thus handled as two completely separate words, shows that the prediction results are consistent. This also indicates that when the count of papers where the term appears is low, the score estimates are less stable, as can be seen with  ``spikevax'' (has a count of 41) and ``modernas'' (count of only 16), which both have lower scores. 
\setlength{\tabcolsep}{3pt}
\begin{table}[!htb]

\footnotesize

    \begin{subtable}{.5\linewidth}
      \caption{ virus strains (``b11529'')}
      \label{tab:b11529}
      \centering
\begin{tabular}{lrrrrr}
\toprule
Word & SCORE & Count & 
Year & Sim  \\ \midrule
b1526         & 0.9999         & 69                         & 2021                               & 0.83                          \\
%b11529        & 0.9991         & 349                        & 2021                               & 1.00                           \\
vocs          & 0.9987         & 810                        & 2004                               & 0.75                           \\
g614          & 0.9984         & 107                        & 2020                               & 0.71                           \\
b16172        & 0.9981         & 714                        & 2021                               & 0.90                           \\
omicron       & 0.9979         & 1,253                       & 2021                               & 0.90                           \\
202012/01     & 0.9977         & 68                         & 2020                               & 0.82                           \\
b1617         & 0.9977         & 122                        & 2021                               & 0.86                          \\
delta         & 0.9973         & 2,873                       & 1996                               & 0.84                          \\
b1429         & 0.9971         & 59                         & 2021                               & 0.79                          \\
b1427         & 0.9967         & 33                         & 2021                               & 0.78                           \\
ba1           & 0.9966         & 91                         & 2020                               & 0.83                           \\
d614g         & 0.9963         & 867                        & 2020                               & 0.80                           \\
501yv2        & 0.9963         & 78                         & 2020                               & 0.84                           \\
ba2           & 0.9959         & 75                         & 2021                              & 0.82                           \\ \bottomrule                               
\end{tabular}
    \end{subtable}%
    \begin{subtable}{.5\linewidth}
      \centering
        \caption{ vaccines (``covaxin'')}
        \label{tab:covaxin}
\begin{tabular}{lrrrrr}
\toprule
Word & SCORE & Count & 
Year & Sim  \\ \midrule
bbv152        & 0.9810         & 53                         & 2020                                & 0.77                            \\
mrna-1273     & 0.9681         & 779                        & 2020                                & 0.72                            \\
%covaxin       & 0.9576         & 95                         & 2021                                 & 1.00                            \\
moderna       & 0.9548         & 890                        & 2020                               & 0.73                            \\
sinovac       & 0.9448         & 109                        & 2020                                & 0.75                            \\
bbibp-corv    & 0.9117         & 81                         & 2020                                 & 0.78                            \\
ncov-19       & 0.8864         & 544                        & 2020                                 & 0.71                            \\
mrna1273      & 0.8818         & 35                         & 2020                                & 0.71                            \\
coronavac     & 0.8767         & 302                        & 2020                                & 0.74                            \\
gam-covid-vac & 0.8715         & 56                         & 2020                                  & 0.77                            \\
sputnik       & 0.8031         & 143                        & 2020                                 & 0.78                            \\
chadox1       & 0.7777         & 767                        & 2013                                 & 0.72                            \\
bnt162b2      & 0.6938         & 2,210                       & 2020                                  & 0.73                            \\
spikevax      & 0.6444         & 41                         & 2021                                 & 0.71                            \\
modernas      & 0.6321         & 16                         & 2020                                  & 0.75                            \\ 
\bottomrule
\end{tabular}
    \end{subtable}

    \begin{subtable}{.5\linewidth}
      \caption{ drugs (``imdevimab'')}
      \label{tab:imdevimab}
      \centering
\begin{tabular}{p{1.9cm}rrrrr}
\toprule
Word & SCORE & Count & 
Year & Sim  \\ \midrule

% imdevimab             & 0.9966         & 86                         & 2021                                 & 1.0                               \\
casirivimab           & 0.9962         & 92                         & 2021                                 & 0.98                               \\
etesevimab            & 0.9929         & 46                         & 2021                                 & 0.91                               \\
bamlanivimab          & 0.9914         & 179                        & 2020                                 & 0.83                               \\
sotrovimab            & 0.9894         & 48                         & 2021                                 & 0.85                               \\
regen-cov             & 0.9882         & 27                         & 2021                                 & 0.84                               \\
ly-cov555             & 0.9669         & 30                         & 2020                                 & 0.81                               \\
casirivimab-imdevimab & 0.9653         & 27                         & 2021                                 & 0.76                               \\
regn-cov2             & 0.9603         & 35                         & 2020                                 & 0.79                               \\
ronapreve             & 0.9278         & 8                          & 2021                                 & 0.74                               \\
regdanvimab           & 0.8993         & 16                         & 2021                                  & 0.71                               \\
monoclonal            & 0.8918         & 3,922                       & 1982                                  & 0.71                               \\
ly-cov016             & 0.8878         & 16                         & 2020                                  & 0.71                               \\
regn10987             & 0.8676         & 19                         & 2020                                  & 0.71                               \\
casirivimab/ imdevimab & 0.8332         & 41                         & 2021                                  & 0.84                               \\ \bottomrule                            
\end{tabular}
    \end{subtable}%
    \begin{subtable}{.5\linewidth}
      \centering
        \caption{similar to ``remdesivir''}
        \label{tab:remdesivir}
\begin{tabular}{lrrrrr}
\toprule
Word & SCORE & Count & 
Year & Sim  \\ \midrule
arbidol             & 0.9996         & 201                        & 2008                  & 0.76                                               \\
molnupiravir        & 0.9996         & 132                        & 2020                 & 0.80                                               \\
ciclesonide         & 0.9934         & 52                         & 2012                 & 0.70                                               \\
nirmatrelvir        & 0.9909         & 27                         & 2022                 & 0.71                                               \\
lpv/r               & 0.9896         & 79                         & 2006                 & 0.74                                               \\
umifenovir          & 0.9876         & 141                        & 2014                 & 0.78                                               \\
lopinavir-ritonavir & 0.9803         & 141                        & 2020                 & 0.82                                               \\
paxlovid            & 0.9756         & 16                         & 2022                 & 0.75                                               \\
lopinavir/ritonavir & 0.9750         & 671                        & 2004                 & 0.85                                               \\
sarilumab           & 0.9741         & 121                        & 2019                 & 0.71                                              \\
ritonavir           & 0.9714         & 331                        & 2004                 & 0.79                                              \\
favipiravir         & 0.9703         & 807                        & 2009                 & 0.89                                               \\
%remdesivir          & 0.9616         & 2,693                      & 2018                 & 1.0                                              \\
nitazoxanide        & 0.9522         & 132                        & 2009                 & 0.79                                              \\
eidd-2801           & 0.9298         & 40                         & 2020                 & 0.75                                              \\                    \\ \\ 
\bottomrule
\end{tabular}
    \end{subtable} 
     \label{tbl:imdrem}     \caption{Top words for selected topics. Sorted by the $\textit{SCORE}(w)$, Sim is the cosine similarity of the word vector to the word vector of the topic in the caption of each table, year indicates the earliest publication year in the corpus when the word appears.
    }
    \label{tbl:b11cov}
\end{table}
\clearpage

\end{document}